\documentclass[letterpaper, 10 pt, conference]{ieeeconf}  

\IEEEoverridecommandlockouts 
\usepackage{cite}
\usepackage{amsmath,amssymb,amsfonts}
  \usepackage{amsthm}

\usepackage{algorithmic}
\usepackage{graphicx}
\usepackage{textcomp}
\usepackage{xcolor}
\usepackage{mathrsfs}
\usepackage{array}
\usepackage{multirow} 
\usepackage{subfigure}
\usepackage{float}
\usepackage{arydshln}
\usepackage[ruled,vlined,lined,boxed,linesnumbered,commentsnumbered]{algorithm2e}
\usepackage{tikz}
\usetikzlibrary{positioning, fit, arrows.meta, shapes.geometric, backgrounds}

\def \R {\mathbb{R}}
\def \P {\mathbb{P}}
\def \bW {\mathbf{W}}
\def \bnu {\boldsymbol{\nu}}
\def \E {\mathbb{E}}
\def \by {\boldsymbol{y}}

\date{}
\title{State estimations and noise identifications with intermittent corrupted observations via Bayesian variational inference}
\author{Peng Sun$^{1}$, Ruoyu Wang$^{2}$, {\it IEEE student member} and Xue Luo$^{3,\sharp}$, {\it IEEE senior member}
\thanks{*This work is financially supported by National Natural Science Foundation of China (Grant No. 12271019) and the National Key R\&D Program of China (Grant No. 2022YFA1005103).}
\thanks{$^{1}$P. Sun is with School of Mathematical Sciences, Beihang University, Beijing, P. R. China 102206.  
        {\tt\small 1512530585@buaa.edu.cn}}%
\thanks{$^{2}$R. Wang is with School of Mathematical Sciences, Beihang University, Beijing, P. R. China 102206. 
        {\tt\small WangRY@buaa.edu.cn}}%
\thanks{$^{3}$X. Luo is with School of Mathematical Sciences, Beihang University, Beijing, P. R. China 102206, and Key Laboratory of Mathematics, Informatics and Behavioral Semantics (LMIB), Beihang University, Beijing, P. R. China 100191.
        {\tt\small xluo@buaa.edu.cn}}%
\thanks{$^{\sharp}$X. Luo is the corresponding author.}
}

\begin{document}
    
 \maketitle
   \begin{abstract}
   This paper focuses on the state estimation problem in distributed sensor networks, where intermittent packet dropouts, corrupted observations, and unknown noise covariances coexist. To tackle this challenge, we formulate the joint estimation of system states, noise parameters, and network reliability as a Bayesian variational inference problem, and propose a novel variational Bayesian adaptive Kalman filter (VB-AKF) to approximate the joint posterior probability densities of the latent parameters. Unlike existing AKF that separately handle missing data and measurement outliers, the proposed VB-AKF adopts a dual-mask generative model with two independent Bernoulli random variables, explicitly characterizing both observable communication losses and latent data authenticity. Additionally, the VB-AKF integrates multiple concurrent multiple observations into the adaptive filtering framework, which significantly enhances statistical identifiability. Comprehensive numerical experiments verify the effectiveness and asymptotic optimality of the proposed method, showing that both parameter identification and state estimation asymptotically converge to the theoretical optimal lower bound with the increase in the number of sensors.
    \end{abstract}
    
    \section{Introduction}

In the field of filtering theory, the Kalman filter (KF) \cite{Kalman1960} is well known to provide optimal state estimation for linear dynamic systems, provided that the exact noise statistics are known a priori. Misspecified noise parameters often degrade filtering performance and may even cause filter divergence. To alleviate this issue, adaptive Kalman filtering (AKF) \cite{1099422} has been developed as an important approach for the joint estimation of system states and unknown noise parameters.
However, in high-dimensional scenarios, an analytical expression for the joint posterior probability density of states and noise parameters is generally unavailable.

Meanwhile, in statistical inference, variational inference (VI) \cite{Blei2016VariationalIA} has been widely adopted to approximate unknown quantities of interest by recasting Bayesian inference into an optimization problem. Compared with traditional Markov Chain Monte Carlo methods, VI exhibits superior computational efficiency when handling complex hierarchical models \cite{BJ:06}. Further improvements in scalability, such as stochastic variational inference \cite{JMLR:v14:hoffman13a}, have greatly extended the applicability of VI to dynamic systems.

    Pioneering this intersection, Särkkä and Nummenmaa \cite{4796261} presented the first Variational Bayesian AKF (VB-AKF) for unknown measurement noise covariance  by approximating the joint probability density functions (pdf) with Gaussian inverse-gamma distributions. This foundational work was subsequently extended to address both unknown process noise covariance and measurement noise covariance in filtering \cite{8025799} and smoothing \cite{7305882} contexts. Building upon these core architectures, researchers have adapted VB to tackle diverse modeling challenges. For instance, Ma et al. \cite{8409977} applied variational Bayesian (VB) to approximate the joint pdf of states and model identities in multiple state-space models, while Xu et al. \cite{9280395} and Xia et al. \cite{9634016} developed VB-based adaptive fixed-lag smoothing and calibration methods, respectively. Most recently, Lan et al. \cite{10818708} advanced the optimization framework itself, proposing a novel AKF method based on conjugate-computation variational optimization to efficiently solve the joint identification problem in complex systems.
    
  Besides, in networked control systems and large-scale wireless sensor networks, intermittent observations (i.e., data missing or packet dropouts) caused by communication channel fading pose a significant challenge. Sinopoli et al. \cite{1333199} first derived the critical divergence threshold of the KF’s estimation covariance under such observations, making this a pivotal research direction with various intermittent observation models explored. For example, Li et al. \cite{Li2015ACC} designed an optimal diffusion KF for distributed sensor networks, while Xu et al. \cite{9123558} used event-triggering to address random delays and observation losses. This problem has since expanded from linear to nonlinear/distributed frameworks, with Kluge et al. \cite{Kluge2010} and Li et al. \cite{Li2012} establishing EKF and UKF stochastic stability under random dropouts, respectively.
  
More recently, research has shifted to adaptive estimation with unknown dropout/noise statistics. A stochastic event-triggered VB filter \cite{9870528} jointly estimates state and unknown noise covariances, and VB-based methods \cite{VB_PacketLoss2024} infer state and measurement loss probability adaptively. However, most existing approaches treat dropouts and corrupted observations separately, lacking a unified framework to simultaneously handle outliers and missing data.

  In this paper, we model intermittent and corrupted observations via two independent Bernoulli random variables, which respectively characterize the packet loss and measurement accuracy of the survival observations. Within this dual-parameter modeling framework, we make the following key contributions: On the one hand, we adapt the VB-AKF to enable effective joint state and noises estimation, as well as the dropout and clean rates, under the scenario of simultaneous packet dropouts and corrupted observations. On the other hand, we propose a centralized sequential fusion scheme for the VB-AKF and validate its statistical properties through  numerical investigations. Specifically, we first demonstrate the asymptotic optimality of the dual-mask framework, revealing that expanding the observation sample size fundamentally drives the inference error to monotonically converge to the theoretical optimal lower bound. We then verify the algorithm's dynamic resilience, confirming its capability of zero-delay trajectory tracking under extreme impulsive interferences, as well as stable variance identification under severe catastrophic scenarios featuring simultaneous massive packet dropouts and data corruption. Finally, comprehensive ablation studies are conducted to rigidly delineate the operational envelope of the proposed method, explicitly characterizing the statistical identifiability boundaries and theoretical inference limits. 

The paper is organized as follows: Section \ref{sec-2} elaborates on the concrete problem to be addressed and provides preliminary background on variational inference. Section \ref{sec-3} derives the VB-AKF tailored to our specific scenario, along with the corresponding pseudo-code. Section \ref{sec-4} presents several numerical experiments to demonstrate the effectiveness of the proposed method. Finally, Section \ref{sec-5} summarizes the key findings and draws the conclusion.

\section{Problem setting-up and Preliminaries}\label{sec-2}

\subsection{Problem setting-up}\label{sec-2.1}
     
       Consider a time series of length $T$ for a linear dynamic system. To model large-scale distributed tracking and network-induced packet dropouts (i.e., missing data), we assume a centralized network consisting of $N$ distributed sensor nodes, all observing a single target trajectory simultaneously. At time $k$, let $x_k\in\R^{d_x}$ denote the system state and $y_{i,k}\in\R^{d_y}$ (for $d_x,d_y\geq1$) denote the observation from the $i$-th sensor. The underlying physical dynamics of the system and the observation model for the $i$-th sensor are given by:
    \begin{align}
        x_k &= F_k x_{k-1} + w_k,\label{eq:state_hd} \\
        y_{i,k} &= \gamma_{i,k}H_k x_k +v_{i,k}
       +(1-z_{i,k})\epsilon_k, \label{eq:obs_hd}
    \end{align}
  for $i=1,\cdots,N$. Here, $F_k \in \mathbb{R}^{d_x \times d_x}$ and $H_k \in \mathbb{R}^{d_y \times d_x}$ represent the state transition matrix and observation matrix, respectively. The system noise $w_k \sim \mathcal{N}(\mathbf{0}, Q_k) $, observation noise $v_{i,k} \sim \mathcal{N}(\mathbf{0}, R_k)$ and extra corrupted noise  $\epsilon_k \sim \mathcal{N}(\mathbf{0}, E_k)$ are mutually independent. We assume that $Q_k \in \mathbb{R}^{d_x \times d_x}$ and $R_k \in \mathbb{R}^{d_y \times d_y}$ are unknown, while $E_k$ is known apriori. 
  
\begin{figure}[htbp]
    \centering
    \resizebox{0.9\linewidth}{!}{
        \begin{tikzpicture}[>=stealth, thick]
            \tikzstyle{var} = [circle, draw, minimum size=0.9cm, inner sep=0pt, fill=white]
            \tikzstyle{obs} = [circle, draw, minimum size=0.9cm, inner sep=0pt, fill=gray!30] 
            \tikzstyle{hyper} = [rectangle, draw, minimum size=0.7cm, inner sep=3pt, fill=white]
            \tikzstyle{plate} = [draw, rectangle, rounded corners, inner sep=0.3cm, dashed, draw=black!60]

            \node[var] (x_prev) at (0, 0) {$x_{k-1}$};
            \node[var] (x_k) at (3, 0) {$x_k$};
            \node[var] (Q_k) at (3, 2.2) {$Q_k$};
            \node[hyper] (hyp_Q) at (0, 2.2) {$\nu_0, V_0$};

            \node[obs] (y_k) at (3, -2.5) {$y_{i,k}$};
            \node[var] (R_k) at (0, -2.5) {$R_k$};
            \node[hyper] (hyp_R) at (-2.5, -2.5) {$u_0, U_0$};
            \node[hyper] (E_k) at (6.5, -2.5) {$E_k$};

            \node[var] (z_k) at (1.5, -4.5) {$z_{i,k}$};
            \node[obs] (gamma_k) at (4.5, -4.5) {$\gamma_{i,k}$};

            \node[var] (beta_k) at (1.5, -6.5) {$\beta_k$};
            \node[var] (rho_k) at (4.5, -6.5) {$\rho_k$};
            \node[hyper] (hyp_beta) at (-1, -6.5) {$a_{\beta,0}, b_{\beta,0}$};
            \node[hyper] (hyp_rho) at (7, -6.5) {$a_{\rho,0}, b_{\rho,0}$};

            \draw[->] (x_prev) -- node[above] {$F_k$} (x_k);
            \draw[->] (Q_k) -- (x_k);
            \draw[->] (hyp_Q) -- (Q_k);

            \draw[->] (x_k) -- node[right] {$H_k$} (y_k);
            \draw[->] (R_k) -- (y_k);
            \draw[->] (hyp_R) -- (R_k);
            \draw[->] (E_k) -- (y_k);

            \draw[->] (z_k) -- (y_k);
            \draw[->] (gamma_k) -- (y_k);

            \draw[->] (beta_k) -- (z_k);
            \draw[->] (rho_k) -- (gamma_k);

            \draw[->] (hyp_beta) -- (beta_k);
            \draw[->] (hyp_rho) -- (rho_k);

            \begin{scope}[on background layer]
                \node[plate, fit=(y_k) (z_k) (gamma_k) , label={[anchor=south east, inner sep=3pt]}] {};
            \end{scope}
        \end{tikzpicture}
    } 
    \caption{Generative model of the linear filtering problem \eqref{eq:state_hd}-\eqref{eq:obs_hd} with packet dropout and corrupted noises. Shaded nodes: observable variables; unshaded circles: latent parameters to be estimated; rectangles: hyper-parameters in priors.}
    \label{fig:pgm}
\end{figure}
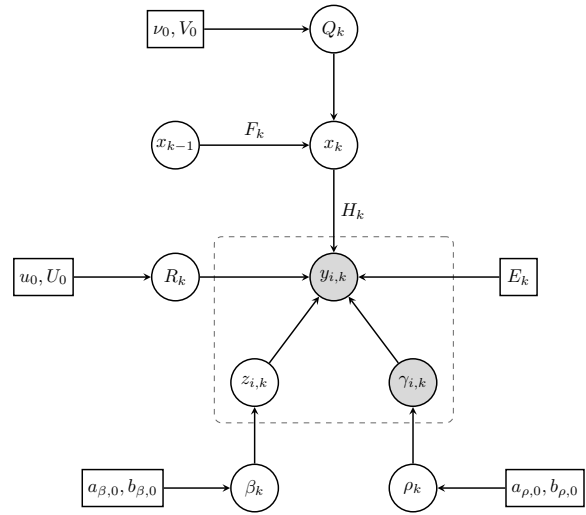

 The binary indicator $\gamma_{i,k}\in\{0,1\}$ characterizes intermittent packet dropouts caused by communication channel fading: specifically, $\gamma_{i,k}=1$ indicates successful reception of the observation from the $i$-th sensor at time $k$, with $\rho_k\in[0,1]$ denoting the network survival rate, i.e. $\P(\gamma_{i,k}=1)=\rho_k$.  Even when observations are successfully received, they may be corrupted by extreme electromagnetic interference or sensor malfunctions, resulting in inaccurate measurements. To characterize the cleanliness of received observations (i.e., whether they are uncorrupted), we introduce another unobservable and independent binary indicator $z_{i,k}\in\{0,1\}$: $z_{i,k}=1$ means the observation from the $i$-th sensor at time $k$ is clean (uncorrupted), whereas $z_{i,k}=0$ indicates a corrupted observation accompanied by the extra noise $\epsilon_k$. The clean rate of the received observations, defined as $\P(z_{i,k}=1)=\beta_k$, is also an unknown parameter to be estimated. The generative model and its known/unknown parameters are displayed in Fig. \ref{fig:pgm}.

To summarize, under the considered problem setup, the following quantities are assumed known apriori: the state transition matrix $F_k$, the observation matrix $H_k$, the extra corruption noise covariance $E_k$, and the noisy/corrupted sensor observations $y_{i,k}$. The binary packet dropout indicator $\gamma_{i,k}$ is observable and also available. The unknown quantities to be jointly estimated include: the system state $x_k$, the unknown noise covariances $Q_k$ and $R_k$, the packet dropout rate $\rho_k$, the clean observation indicator $z_{i,k}$, and the clean rate $\beta_k$.

    \subsection{Preliminary: Variational Inference (VI)}
    
    Let us denote the latent parameter to be estimated as $\bW=(W_1,\cdots,W_n)$ and the observed data as $Z$. VI aims to find a tractable distribution family $q(\bW)$ to approximate the complex true posterior distribution $p(\bW|Z)$ \cite{Blei2016VariationalIA, BJ:06}. This goal can be achieved by maximizing the well-known Evidence Lower Bound (ELBO):
    \begin{equation*}
        \mathcal{L}(q):= \mathbb{E}_{q(\bW)}[\log p(Z, \bW)] - \mathbb{E}_{q(\bW)}[\log q(\bW)].
    \end{equation*}

     Assume that the true conditional distribution belongs to the exponential family:
     \begin{equation}\label{eqn-2.1}
        p(W_i | W_{-i}, Z) = h(W_i) \exp\left\{ g_i^T(\bW_{-i}, Z) W_i - A(g_i) \right\},
    \end{equation}
    where $g_i$ is the natural parameter depending on the rest of the variables $\bW_{-i}$ and the observed data $Z$, and $A(\cdot)$ is the log-partition function. Furthermore, with the mean-field assumption each component of $\bW$ is assumed to be independent parameters, and we restrict the variational distribution $q(W_i)$ to the same exponential family \eqref{eqn-2.1} with a free natural parameter $\nu_i$, i.e.
    \[q_{\bnu}(\bW) = \prod_{i=1}^n q_{\nu_i}(W_i),\]
    with $\bnu=(\nu_1,\cdots,\nu_n)$ and         \begin{equation*}
        q_{\nu_i}(W_i) = h(W_i) \exp\left\{ \nu_i^T W_i - A(\nu_i) \right\}.
    \end{equation*}
    The local ELBO with respect to $W_i$ is then defined to be 
    \begin{equation}\label{eqn-2.2}
        \mathcal{L}_i(q_{\nu_i}):= \mathbb{E}_{q_{\nu_i}}[\log p(W_i | \bW_{-i}, Z)] - \mathbb{E}_{q_{\nu_i}}[\log q_{\nu_i}(W_i)].
    \end{equation}

    It is the key fact \cite{BJ:06} that by minimizing the local ELBO \eqref{eqn-2.2} the optimal natural parameter $\nu_i$ equals to the expectation of $g_i$ in the true conditional distribution:
    \begin{equation} \label{eq:vi_basic}
        \nu_i^* = \mathbb{E}_{q_{\bnu}}[g_i(\bW_{-i}, Z)].
    \end{equation} 
    For a detailed derivation of \eqref{eq:vi_basic}, interested readers are referred to Appendix A in \cite{BJ:06}. Equation \eqref{eq:vi_basic} plays a critical role in the subsequent parameter estimation procedures.

    \section{Structured VI at time instant $k$}\label{sec-3}

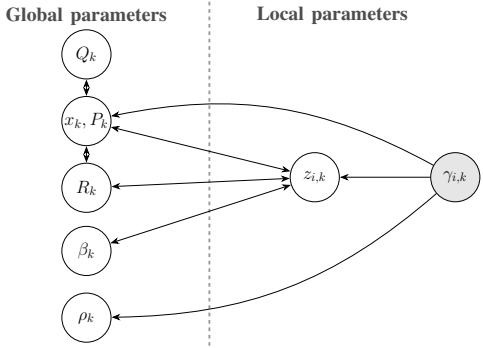
\begin{figure}[htbp]
    \centering
    \resizebox{0.75\linewidth}{!}{
        \begin{tikzpicture}[>={Stealth[scale=1.1]}, thick]
            \tikzstyle{var} = [circle, draw, minimum size=1.4cm, inner sep=0pt, fill=white, font=\Large]
            \tikzstyle{obs} = [circle, draw, minimum size=1.4cm, inner sep=0pt, fill=gray!20, font=\Large]

            \draw[dashed, black!40, line width=1.5pt] (0, 5.0) -- (0, -4.8);
            
            \node[font=\LARGE\bfseries, text=black!70] at (-3.5, 4.6) {Global parameters};
            \node[font=\LARGE\bfseries, text=black!70] at (3.5, 4.6) {Local parameters};

            
            \node[var] (Q)    at (-3.5, 3.5)  {$Q_k$};
            \node[var] (x)    at (-3.5, 1.6)  {$x_k,P_k$};
            \node[var] (R)    at (-3.5, -0.3)    {$R_k$};
            \node[var] (beta) at (-3.5, -2.1) {$\beta_k$};
            \node[var] (rho)  at (-3.5, -4) {$\rho_k$};

            \node[var] (pi)    at (3.0, 0)    {$z_{i,k}$};
            \node[obs] (gamma) at (7.0, 0)    {$\gamma_{i,k}$};

            
            \draw[<->] (x) -- (Q);
            \draw[<->] (x) -- (R);

            \draw[<->] (x) -- (pi);
            \draw[<->] (R) -- (pi);
            \draw[<->] (beta) -- (pi);

            \draw[->] (gamma) -- (pi);
            
            \draw[->, bend right=20] (gamma) to (x);
            \draw[->, bend left=20]  (gamma) to (rho);

        \end{tikzpicture}
    } 
    \caption{Latent parameters dependency. Global parameters shared across all sensor nodes, while local parameters (subscript $i,k$) are different among individual sensors.}
    \label{fig:simple_vi_dep}
\end{figure}

Recalling the problem setup described in Section \ref{sec-2.1}, we illustrate the parameters to be estimated and their mutual relationships in Fig. \ref{fig:simple_vi_dep}. We refer to the parameters shared by all sensors as global parameters, and those intrinsic to each individual sensor as local parameters.

The latent parameter in our problem is $\bW_{k} :=\{x_k, Q_k, R_k, \rho_k, \beta_k, z_{i,k}, i=1,\cdots,N\}$. Under the mean-field assumption \cite{JMLR:v14:hoffman13a}, the variational distribution 
    \begin{align}\label{eq:mfvi}\notag
    q(\bW_k) 
     = &q(x_k) q_{\nu_{k}, V_{k}}(Q_k) q_{u_{k}, U_{k}}(R_k) \\
     &\cdot q_{a_{\rho,k}, b_{\rho,k}}(\rho_k) q_{a_{\beta,k}, b_{\beta,k}}(\beta_k)\prod_{i=1}^N q_{\pi_{i,k}}(z_{i,k}).
    \end{align}
    
To perform the Bayesian inference and to ensure the posterior distribution of the parameters remains tractable, we assign conjugate priors to all parameters. Specifically, inverse-Wishart ($\mathcal{IW}$) priors are adopted for the noise covariance matrices, and Beta priors are imposed on the global network survival rate $\rho_k \in [0,1]$ and the data clean rate $\beta_k \in [0,1]$, i.e.
      \begin{align}\label{eq:prior_Q}     `
            Q_k &\sim \mathcal{IW}(\nu_0, V_0),\\\label{eq:prior_R}
            R_k &\sim \mathcal{IW}(u_0, U_0),  \\\label{eq:prior_rho}
            \rho_k &\sim \text{Beta}(a_{\rho,0}, b_{\rho,0}), \\\label{eq:prior_beta}
            \beta_k&\sim \text{Beta}(a_{\beta,0}, b_{\beta,0}).
      \end{align}

The likelihood function for the i-th sensor forms a Bayesian mixture of Gaussians (BMG) \cite{Blei2016VariationalIA}, which is ``gated" by the packet dropout indicator $\gamma_{i,k}$. Specifically, when  $\gamma_{i,k}=1$ (i.e., the observation is successfully received), the likelihood is given by:
    \begin{align}\label{eq:obs_gmm}
    &p(y_{i,k} \mid x_k, R_k, z_{i,k}, \gamma_{i,k}=1)\\\notag
    =&\underbrace{\mathcal{N}(y_{i,k} \mid H_k x_k, R_k)^{z_{i,k}}}_{\text{Clean observation}} \cdot \underbrace{\mathcal{N}(y_{i,k} \mid H_k x_k, R_k + E_k)^{1-z_{i,k}}}_{\text{Corrupted observation}};
    \end{align}
    whereas if $\gamma_{i,k}=0$ (i.e., packet dropout occurs), the observation $y_{i,k}$ is unavailable.
    

    \subsection{Bayesian Mixture of Gaussians (BMG) and inference of clean rate}\label{sec-3.1}
  

   Let us infer the latent clean indicator $z_{i,k}$ for each $i$-th sensor by BMG proposed in \cite{Blei2016VariationalIA}. From the likelihood \eqref{eq:obs_gmm} and the prior $\P(z_{i,k}=1)=\beta_k$, the conditional log-probability for the clean observation ($z_{i,k}=1$) is given by:
    \begin{equation*}
        \ln p(z_{i,k}=1 | \bW_{-z_{i,k}}, y_{i,k}) = \ln \beta_k + \ln \mathcal{N}(y_{i,k} | H_k x_k, R_k),
    \end{equation*}
    up to a constant. According to \eqref{eq:vi_basic}, taking the expectation with respect to $q(x_k, R_k, \beta_k)$, we obtain the variational log-posterior:
    \begin{align} \label{eq:log_q_z1}\notag
      &  \ln q^*(z_{i,k}=1) \\\notag
      =& \mathbb{E}_{q(\boldsymbol{x}, R_k, \beta_k)} \Bigg[ \ln \beta_k - \frac{1}{2}\ln|R_k| \\\notag
        &\phantom{aaaaaaaaa}- \frac{1}{2} (y_{i,k} - H_k x_k)^T R_k^{-1} (y_{i,k} - H_k x_k) \Bigg]\\
        &+\textup{const}.
    \end{align}
 Let us approximate $q(x_k)\approx \mathcal{N}(\hat{x}_{k|k-1}, P_{k|k-1})$, where 
     \begin{align*}
                \hat{x}_{k|k-1} =& F_k \hat{x}_{k-1|k-1},\\
                P_{k|k-1}=& F_k P_{k-1|k-1} F_k^T + (\E Q_k)^{-1},
     \end{align*}
     are the predictive mean and covariance matrix. The expectation of the quadratic term in \eqref{eq:log_q_z1} can be obtained analytically:
    \begin{align}\label{eq:Delta1}
        &\mathbb{E}_{q(x_k, R_k)} \left[ (y_{i,k} - H_k x_k)^T R_k^{-1} (y_{i,k} - H_k x_k) \right] \nonumber \\
        &= (y_{i,k} - H_k \hat{x}_{k|k-1})^T \mathbb{E}[R_k^{-1}] (y_{i,k} - H_k \hat{x}_{k|k-1}) \nonumber \\
        &\quad + \text{Tr}\left(\mathbb{E}[R_k^{-1}] H_k P_{k|k-1} H_k^T\right).
    \end{align}
  Substituting \eqref{eq:Delta1} back to \eqref{eq:log_q_z1}, one has
   \begin{align} \label{eq:rho1} \notag
        &\ln q^*(z_{i,k}=1) \\\notag
        =& \mathbb{E}_{q(\beta_k)}\ln \beta_k - \frac{1}{2}\mathbb{E}_{q(R_k)}\ln|R_k|\\\notag
        &-\frac12(y_{i,k} - H_k \hat{x}_{k|k-1})^T \mathbb{E}[R_k^{-1}] (y_{i,k} - H_k \hat{x}_{k|k-1})\\
        &-\frac12 \text{Tr}\left(\mathbb{E}[R_k^{-1}] H_k P_{k|k-1} H_k^T\right)=:\Delta_{i,k}^1,
    \end{align}
    up to a constant. Similarly, one has
    \begin{align}\label{eq:rho0}\notag
        &\ln q^*(z_{i,k}=0)\\\notag
        =& \mathbb{E}_{q(\beta_k)}\ln (1-\beta_k) - \frac{1}{2}\mathbb{E}_{q(R_k)}\ln|R_k+E_k|\\\notag
        &- \frac{1}{2}(y_{i,k} - H_k \hat{x}_{k|k-1})^T \mathbb{E}(R_k+E_k)^{-1} (y_{i,k} - H_k \hat{x}_{k|k-1})\\
        &-\frac12 \text{Tr}\left(\mathbb{E}(R_k+E_k)^{-1} H_k P_{k|k-1} H_k^T\right)=:\Delta_{i,k}^0,
    \end{align}
    up to a constant. Therefore, the optimal posterior distribution of $z_{i,k}$ is obtained via the softmax transformation:
    \begin{equation}\label{eq:pi_softmax}
        \pi_{i,k} = q^*(z_{i,k}=1)\overset{\eqref{eq:rho1},\eqref{eq:rho0}}= \frac{\Delta_{i,k}^1}{\Delta_{i,k}^1+ \Delta_{i,k}^0}.
    \end{equation}

    \subsection{Inferences of global parameters ($\rho_k$, $\beta_k$, $R_k$ and $Q_k$)}
    \subsubsection{Rates inference $\rho_k$ and $\beta_k$}
    From Fig. \ref{fig:simple_vi_dep}, it is clear to see that $\rho_k$ and $\beta_k$ are only related to two independent Boolean indicators $\gamma_{i,k}$ and $z_{i,k}$, respectively. According to Bayes' rule, one has
    \begin{align*}
       & p(\rho_k | \bW_{-\rho_k}, \boldsymbol{y}) \propto p(\rho_k) \prod_{i=1}^N p(\gamma_{i,k} | \rho_k), \\
        \overset{\eqref{eq:prior_rho}}=& \text{Beta}\Big(a_{\rho,0} + \sum_{i=1}^N \gamma_{i,k}, b_{\rho,0} + \sum_{i=1}^N (1-\gamma_{i,k})\Big),
    \end{align*}
    and
    \begin{align*}
        &p(\beta_k | \bW_{-\beta_k}, \boldsymbol{y}) 
        \propto p(\beta_k) \prod_{i=1}^N p(z_{i,k} | \beta_k)^{\gamma_{i,k}}\\
        &\overset{\eqref{eq:prior_beta}}= \text{Beta}\Big(a_{\beta,0} + \sum_{i=1}^N \gamma_{i,k} \pi_{i,k},  b_{\beta,0} + \sum_{i=1}^N \gamma_{i,k} (1-\pi_{i,k})\Big),
    \end{align*}
    where $\by=\{y_{i,k}\}$, $i=1,\cdots,N$, $k=1,\cdots,T$.

    \subsubsection{Observation covariance matrix $R_k$}
    From Fig. \ref{fig:simple_vi_dep}, the observation covariance matrix $R_k$ depends on $z_{i,k},\gamma_{i,k}$ and $x_k$. According to Bayes' rule, we have
      \begin{align}\label{eqn-3.3}\notag
        &\ln p(R_k | \bW_{-R_k}, \boldsymbol{y}) \\
        =& \ln p(R_k)+ \sum_{i=1}^N \gamma_{i,k} z_{i,k} \ln \mathcal{N}(y_{i,k} | H_k x_k, R_k) \\\notag
        &+\sum_{i=1}^N \gamma_{i,k}(1- z_{i,k}) \ln \mathcal{N}(y_{i,k} | H_k x_k, R_k+E_k)+ \textup{const}.
    \end{align}
With the prior distribution \eqref{eq:prior_R}, adhering to \eqref{eqn-3.3} would render the posterior distribution of $R_k$ intractable. To address this, we approximate \eqref{eqn-3.3} by omitting the term corresponding to $z_{i,k}=1$. Under this approximation, the posterior distribution of $R_k$ reduces to an inverse-Wishart distribution with $u_{k} = u_0 + \sum_{i=1}^N \gamma_{i,k} \pi_{i,k}$ and the scale matrix is gated by $\pi_{i,k}$, $z_{i,k}$ and $\gamma_{i,k}$, i.e.
    \begin{align}
        U_{k} &= U_0 + \sum_{i=1}^N \gamma_{i,k} \pi_{i,k} \Big[ (y_{i,k} - H_k \hat{x}_{k|k})(y_{i,k} - H_k \hat{x}_{k|k})^T \nonumber \\
        &\quad\quad\quad + H_k P_{k|k} H_k^T \Big]. \label{eq:UN_isolated}
    \end{align}
    
    \subsubsection{State covariance matrix $Q_k$}
     From Fig. \ref{fig:simple_vi_dep}, the state covariance matrix $Q_k$ depends only on $x_k$. According to Bayes' rule, one has
    \begin{align*}
        &\ln p(Q_k |\bW_{-Q_k}, \boldsymbol{y})\\
        =& \ln p(Q_k) + \ln \mathcal{N}(x_k | F_k x_{k-1}, Q_k) + \textup{const}.
    \end{align*}
 Therefore, with the prior distribution \eqref{eq:prior_Q}, the posterior distribution of $Q_k$ is still an inverse-Wishart distribution $\mathcal{IW}(\nu_k,V_k)$, with $\nu_{k} = \nu_0 + 1$ and
    \begin{align}\label{eq:VN}
        V_{k} &= V_0 + {(\hat{x}_{k|k} - F_k \hat{x}_{k-1|k-1})(\hat{x}_{k|k} - F_k \hat{x}_{k-1|k-1})^T} \\\notag
        &\quad + {P_{k|k} + F_k P_{k-1|k-1} F_k^T}-{(F_k P_{k,k-1}^T + P_{k,k-1} F_k^T)}, 
    \end{align}
    where $P_{k,k-1}=P_{k|k}P_{k|k-1}^{-1}F_kP_{k-1|k-1}$ is the Rauch-Tung-Striebel smoother \cite{rauch1965maximum}.

    \subsection{Gated Kalman gain and state's fusion}\label{sec-3.4}
    
    To fuse information from the distributed sensor network seamlessly, we propose a sequential gated updating for the posterior state $\hat x_{k|k}$ and the covariance matrix $P_{k|k}$. For the $i$-th sensor at time $k$, the intermediate state and covariance be $\hat{x}_{k|k}^{[i]}$ and $P_{k|k}^{[i]}$ are updated according to 
     \begin{align}\label{eqn-3.2}
        \hat{x}_{k|k}^{[i]} &= \hat{x}_{k|k}^{[i-1]} + K_{i,k} \left(y_{i,k} - H_k \hat{x}_{k|k}^{[i-1]}\right), \\
        P_{k|k}^{[i]} &= (I - K_{i,k} H_k) P_{k|k}^{[i-1]},
    \end{align}
    where $\hat x_{k|k}^{[i-1]}$ and $P_{k|k}^{[i-1]}$ are taking the place of the predictive state  $\hat x_{k|k-1}$  and covariance matrix $P_{k|k-1}$ in the classical KF. Also, the localized gated gain is constructed as:
    \begin{equation}
        K_{i,k}= \gamma_{i,k} P_{k|k}^{[i-1]} H_k^T \Big( (\Omega_{i,k})^{-1} + H_k P_{k|k}^{[i-1]} H_k^T \Big)^{-1}, \label{eq:gated_gain}
    \end{equation}
    with the measurement precision formulated as a linear combination of the clean and corrupted precision matrices:
    \begin{equation}\label{eqn-3.1}
        \Omega_{i,k} = \pi_{i,k} \mathbb{E}R_k^{-1}+ (1-\pi_{i,k}) \mathbb{E}(R_k+E_k)^{-1}.
    \end{equation}
   
    Since the log-likelihood term is explicitly modulated by $\gamma_{i,k}$, when a network dropout occurs at the $i$-th sensor at time $k$ (i.e. $\gamma_{i,k}=0$), the localized Kalman gain collapses to a zero matrix \cite{1333199}. In this case, the intermediate state estimate and error covariance at the $i$-th sensor automatically degenerate to the previous sensor's intermediate state and covariance matrix.

    \subsection{Iteration of variational inference (VI) and pseudo code }

   The VI steps detailed in Sections \ref{sec-3.1}–\ref{sec-3.4} are implemented iteratively at each time instant to mitigate the bias induced by the parameter priors. The iteration process is indexed by the superscript j in Algorithm \ref{alg:vb_akf}. Meanwhile, the overall procedure of the proposed VB-AKF, which integrates the BMG and Beta-Bernoulli gating mechanisms, is summarized in Algorithm \ref{alg:vb_akf}.
   
\begin{algorithm}[h!]
	\caption{Our proposed VB-AKF}	\label{alg:vb_akf}
	\KwIn{The number of sensors: N; the number of time steps: T; all the Observations $\by^{N\times T}$; the Boolean dropout indicator matrix $\boldsymbol{\Gamma}=(\gamma_{i,k})_{i=1,\cdots,N, k=1,\cdots,T} \in \{0,1\}^{N \times T}$; the number of iterations $J$; hyper-parameters $E_k,a_{\rho,0},b_{\rho,0},a_{\beta,0},b_{\beta,0},u_0,U_0,\nu_0,V_0$ and the initial state $x_0\sim \mathcal{N}(\mu_0,\Sigma_0)$.}
	\BlankLine
	
    \For{k=1:T}{
        \% Dropout rate inference ($\rho_k$)
        
        $\mathbb{E}[1-\rho_k] = \frac{b_{\rho,0} + N - M_k}{a_{\rho,0} + b_{\rho,0} + N}$, where $M_k = \sum_{i=1}^N \gamma_{i,k}$.
        
        \% Iterations of VI at each time instant 
        
        \For{j=1:J}{
            \% Prediction
            \begin{align*}
                \hat{x}_{k|k-1}^{(j)} =& F_k \hat{x}_{k-1|k-1},\\
                P_{k|k-1}^{(j)} =& F_k P_{k-1|k-1} F_k^T + (\E{Q_k}^{(j-1)})^{-1}.
            \end{align*}
            \% Initialize fusion: $\hat{x}_{k|k}^{[0]} = \hat{x}_{k|k-1}^{(j)}$ and $P_{k|k}^{[0]} = P_{k|k-1}^{(j)}$.
            
            \% Local parameter inference \& state and covariance fusion
            
            \For{i=1:N}{
                The clean rate $\pi_{i,k}^{(j)}$ is inferred by \eqref{eq:pi_softmax}.\\
                Calculate equivalent precision $\Omega_{i,k}^{(j)}$ and the local Kalman gain $K_{i,k}^{(j)}$ via  \eqref{eq:gated_gain} and \eqref{eqn-3.1}.\\
                Sequential state and covariance update: $\hat{x}_{k|k}^{[i]\,(j)}, P_{k|k}^{[i]\,(j)}$ via \eqref{eqn-3.2}.
            }
            State and covariance fusion: $\hat{x}^{(j)}_{k|k} = \hat{x}_{k|k}^{[N]\,(j)}$ and $P_{k|k}^{(j)} = P_{k|k}^{[N]\,(j)}$.

            \% Inference of global parameters
            
            Corrupted rate inference: $\mathbb{E}[1-\beta_k^{(j)}] = \frac{b_{\beta,0} + \sum_{i} \gamma_{i,k}(1-\pi_{i,k}^{(j)})}{a_{\beta,0} + b_{\beta,0} + M_k}$.\\
            Update expected precisions:
            \begin{align*}
                \E R_k^{(j)\,-1} =& \Big(u_0 + \sum_{i=1}^N \gamma_{i,k}\pi_{i,k}^{(j)}\Big) U_{k}^{(j)\,-1},\\
                \E Q_k^{(j)\,-1} =& (\nu_0 + 1) V_{k}^{(j)\,-1},
            \end{align*}
            where $U_{k}^{(j)}$ and $V_{k}^{(j)}$ are via \eqref{eq:UN_isolated} and \eqref{eq:VN}, respectively.
        }
        \KwOut{Estimates $\hat{x}_{k|k}=\hat x_{k|k}^{(J)}$, $P_{k|k}=P_{k|k}^{(J)}$; dropout rate $\E\left(1-\rho_k^{(J)}\right)$; corrupted rate $\E\left(1-\beta_k^{(J)}\right)$; noise covariance matrices $\mathbb{E}Q_k^{(J)\,-1}$ and $\mathbb{E}R_k^{(J)\,-1}$.}
    }
\end{algorithm}

\section{Numerical experiments}\label{sec-4}

To comprehensively evaluate the statistical properties and estimation performance of the proposed VB-AKF, we design four numerical experiments. These experiments aim to verify the following characteristics: (1) Asymptotic optimality: To validate that increasing the number of observation sample paths reduces the inference error toward the theoretical lower bound. (2) Transient robustness: To evaluate the tracking ability and covariance recovery performance of the filter under sudden noise variance spikes. (3) Inference under severe degradation: To verify the consistent parameter identifiability under simultaneous high-rate packet dropouts and measurement corruptions. (4) Sensitivity analysis: To reveal the statistical identifiability boundaries and theoretical performance limits of the model via comprehensive ablation studies.

For visualization clarity, we set the state and observation dimensions to $d_x=d_y=1$ to intuitively trace the variance compensation dynamics. In all experiments, we set $F=H=1$, $T=120$ and set the number of VI iterations to $J=20$. Packet dropouts and corrupted measurements are not considered in the first two experiments.

\subsection{Asymptotic optimality}\label{sec-4.1}

In this experiment, we investigate the impact of the observation sample size $N=M_k$ (no packet dropout is considered)  on our proposed VB-AKF's convergence accuracy for linear filtering problem \eqref{eq:state_hd}-\eqref{eq:obs_hd}. The true baseline variances are set to $Q=0.1$ and $R=1$. We compare the Root Mean Square Error (RMSE) of the state in the standard KF, which serves as an ``Oracle" (i.e., it has full knowledge of the true values of $Q$ and $R$), with that of the VB-AKF, which relies entirely on blind posterior estimation. We conduct the experiment by iterating over different values of $N$, specifically $N \in \{1, 2, 5, 10, 20, 50, 100\}$.

\begin{figure}[htbp]
    \centering
    \includegraphics[width=1.0\linewidth]{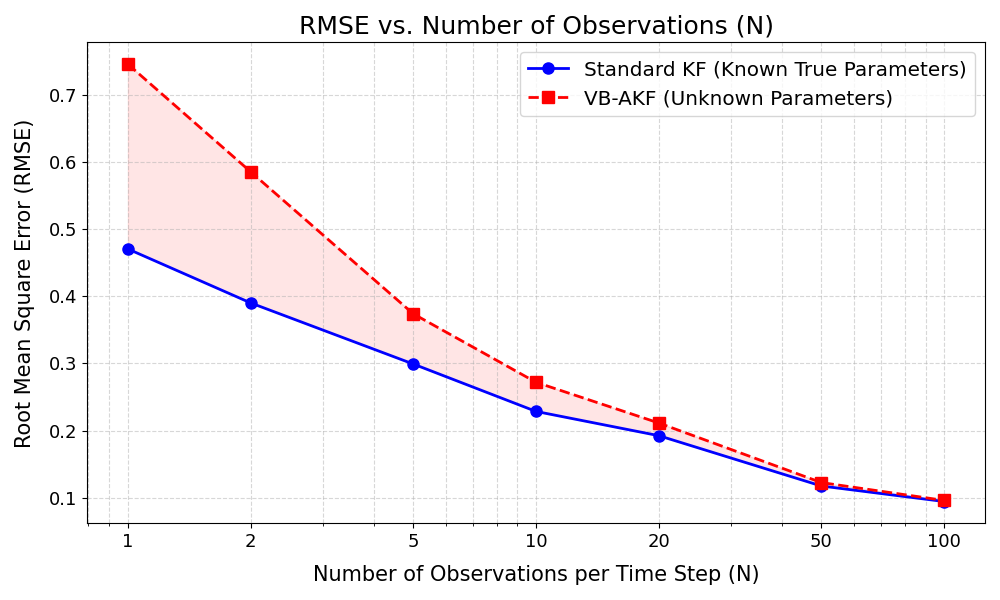}
    \caption{The RMSE of the state convergence comparison between the oracle KF and the proposed VB-AKF with different numbers of observation nodes $N$.}
    \label{fig:exp_m_convergence}
\end{figure}

As shown in Fig.~\ref{fig:exp_m_convergence}, in the data-scarce regime (e.g., $N\leq5$), the RMSE of the VB-AKF is slightly higher than that of the Oracle KF. However, with the accumulation of observation data fusion ($N\geq10$), the error curve of the VB-AKF decreases rapidly and converges closely to the theoretical optimal lower bound. This result numerically validates the core statistical property of Bayesian inference: as the amount of data evidence increases, the adaptive learning mechanism of the algorithm mitigates prior uncertainty, enabling the filtering system to attain the theoretical asymptotic optimality.

\subsection{Transient robustness}\label{sec-4.2}

In the second experiment, we introduce extreme nonstationary variance shifts, with a fixed number of observations $N=M_k=5$ at each time step (packet dropouts are not considered). The baseline variances are set to $Q=0.1$ and $R=1$, consistent with Section \ref{sec-4.1}. Two abrupt anomalies are introduced: a sudden process variance spike at $k=40$ (the process noise variance instantly jumps to $30$), and a simultaneous observation disturbance at $k=80$ (the observation noise variance instantly surges to $60$).

\begin{figure}[htbp]
    \centering
    \includegraphics[width=1.0\linewidth]{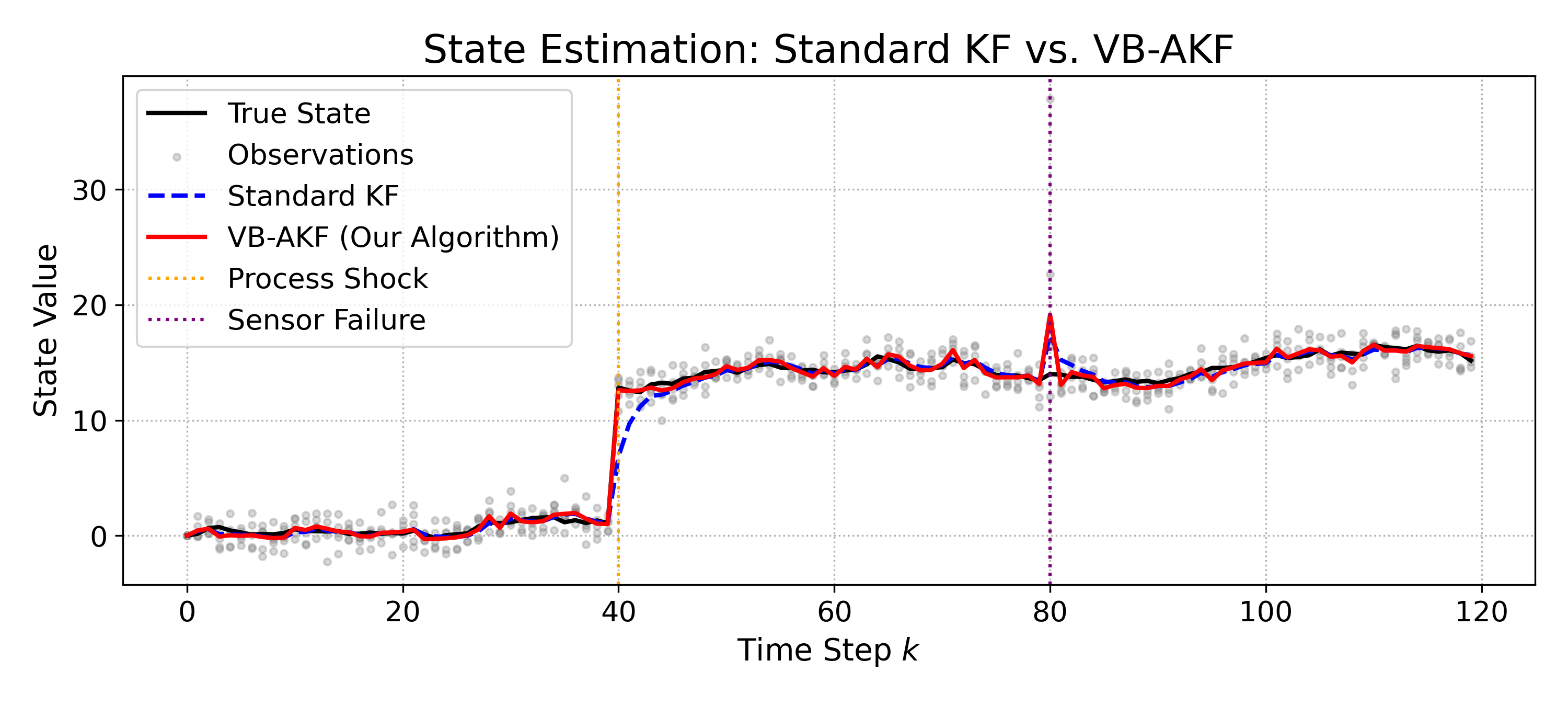} \\
    \vspace{2mm}
    \includegraphics[width=1.0\linewidth]{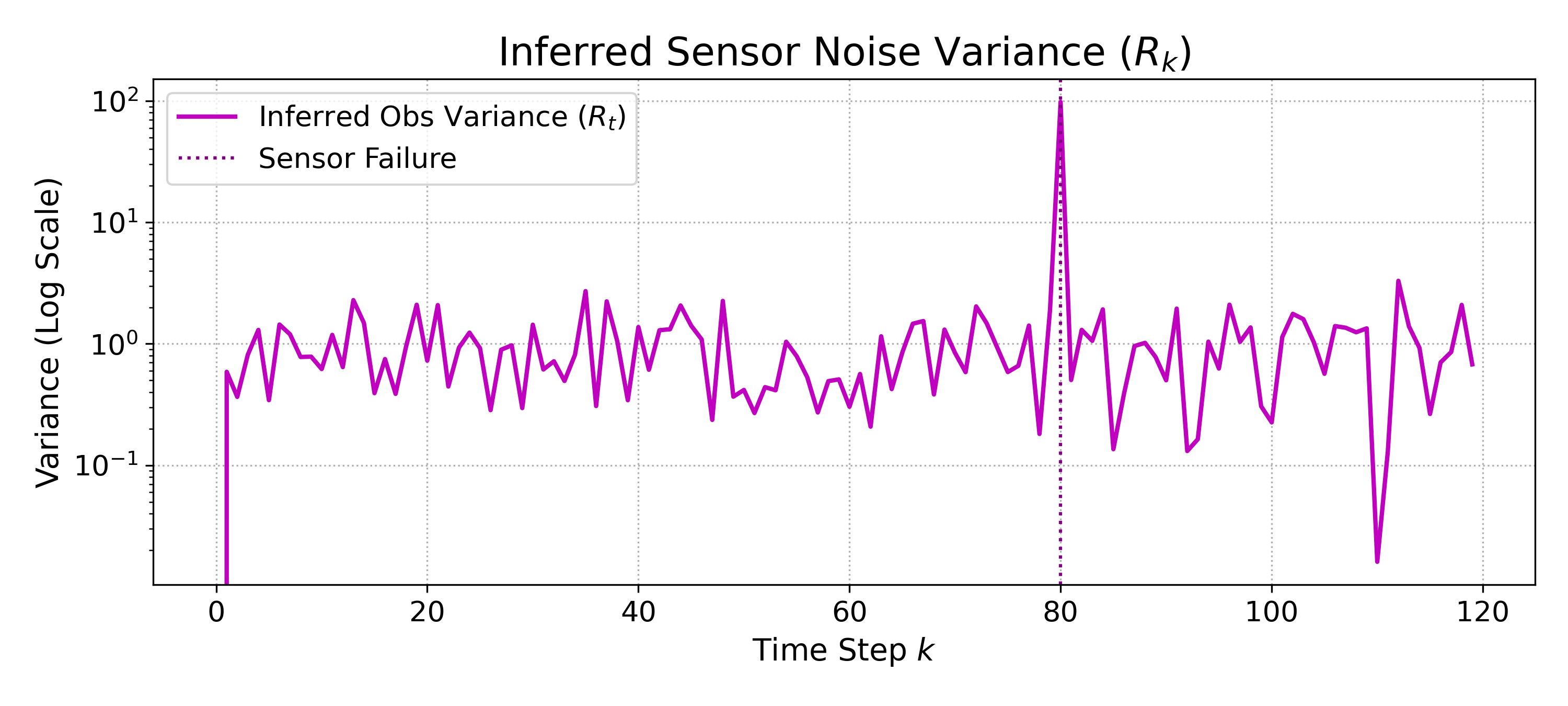} \\
    \vspace{2mm}
    \includegraphics[width=1.0\linewidth]{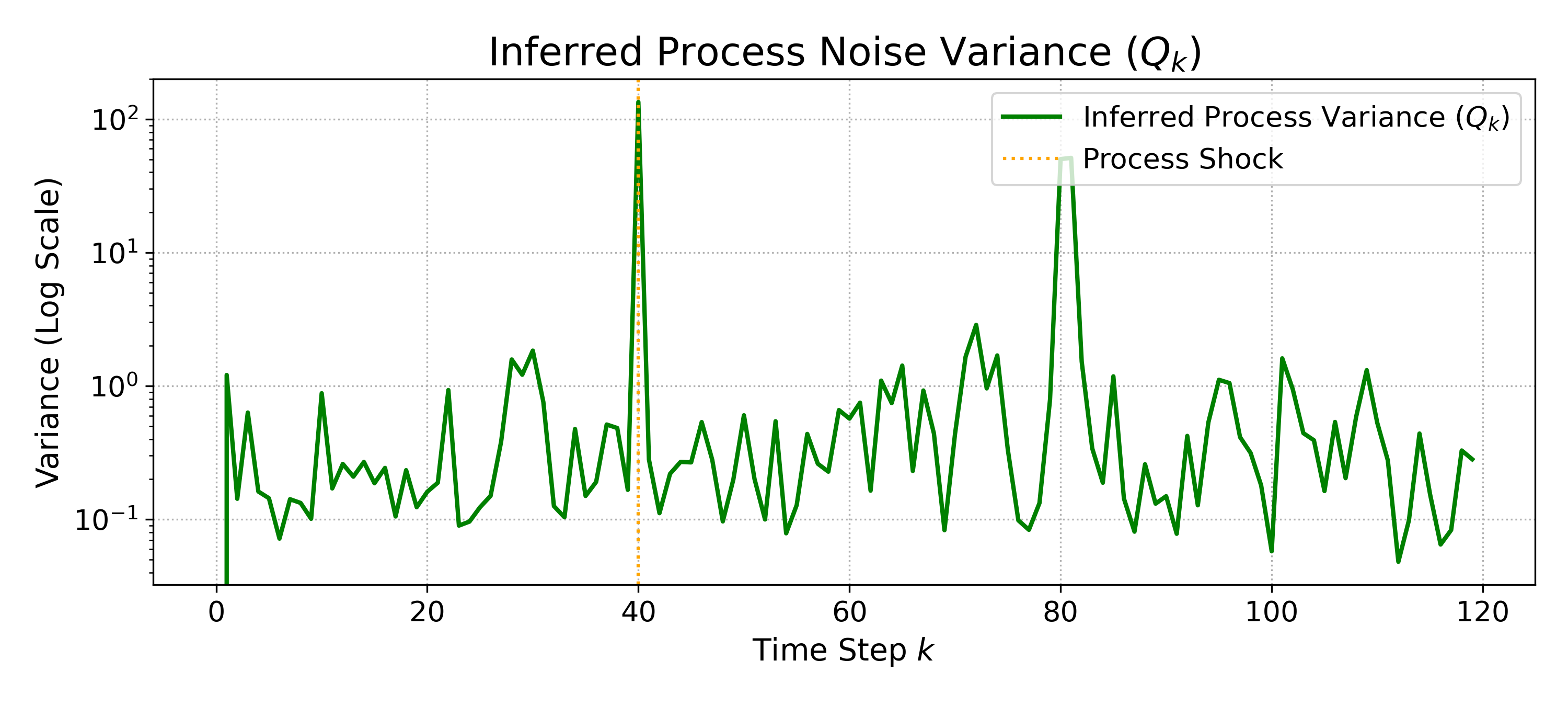}
    \caption{State estimation and noise variance inferences under non-stationary perturbations with $N=5$ sensor nodes. Top: State trajectory tracking; Middle: Inference of observation noise variance $R_k$; Bottom: Inference of process noise variance $Q_k$.}
    \label{fig:exp_m5}
\end{figure}

As shown in Fig.~\ref{fig:exp_m5}, constrained by the static variance assumption, the standard KF responds sluggishly to dynamic maneuvers and is severely degraded by abrupt outliers. In contrast, by performing joint inference across multiple distributed observations, the proposed VB-AKF can rapidly capture the underlying nonstationary variance shifts, thus effectively suppressing invalid perturbations and achieving nearly zero-delay state tracking. Notably, a sudden increase in $R_k$ noticeably affects the estimation of $Q_k$, whereas an increase in $Q_k$ has only a mild influence on the estimation of $R_k$, revealing an asymmetric coupling structure between the process and observation noise covariances during adaptive inference. Moreover, a sudden surge in the observation noise variance degrades the state estimation performance more severely than an surge in the process noise, for both the standard KF and the proposed VB-AKF.

\subsection{Inference under severe degradation}\label{sec-4.3}

To rigorously verify the centralized sequential fusion architecture and the statistical identifiability of the proposed VB-AKF, we consider a network with $N=200$ distributed sensor nodes monitoring the same state process. The true noise covariances are set to $Q=0.05$ and $R=1$, respectively. To simulate an extreme sensing environment, the anomaly perturbation is set to $E_k=10$. A severe data degradation stage is introduced during $k\in[50,100]$, during which the packet dropout rate abruptly rises to 60\%, while the data corruption rate for successfully received packets simultaneously increases to 60\%.

\begin{figure}[htbp]
    \centering
    \includegraphics[width=1.0\linewidth]{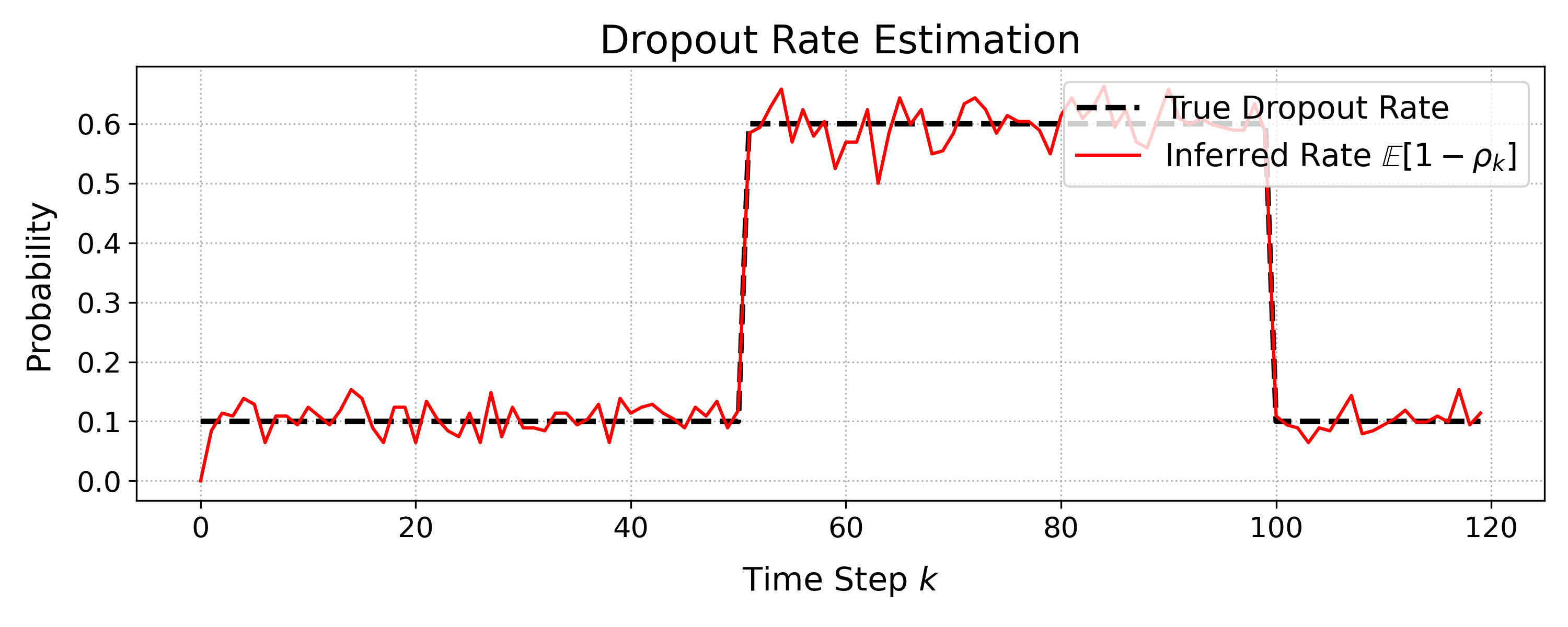} \\
    \vspace{1.5mm}
    \includegraphics[width=1.0\linewidth]{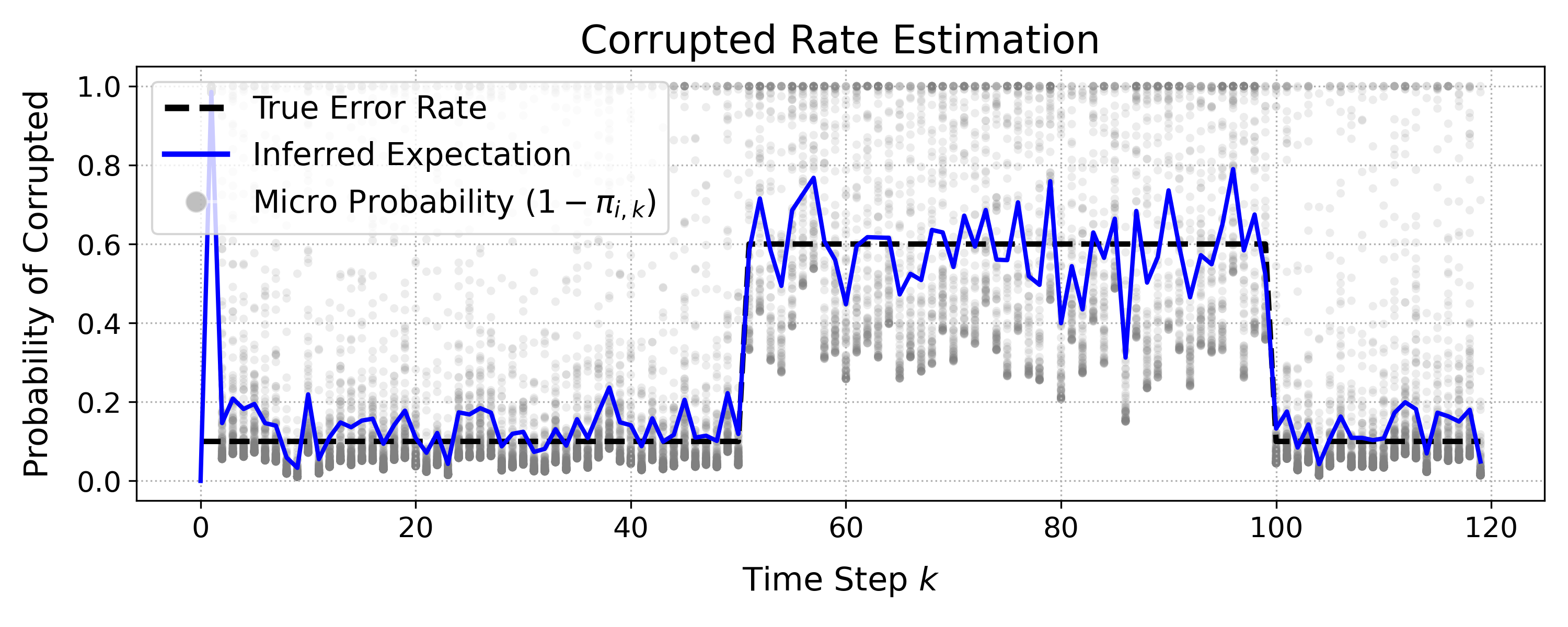} \\
    \vspace{1.5mm}
     \includegraphics[width=1.0\linewidth]{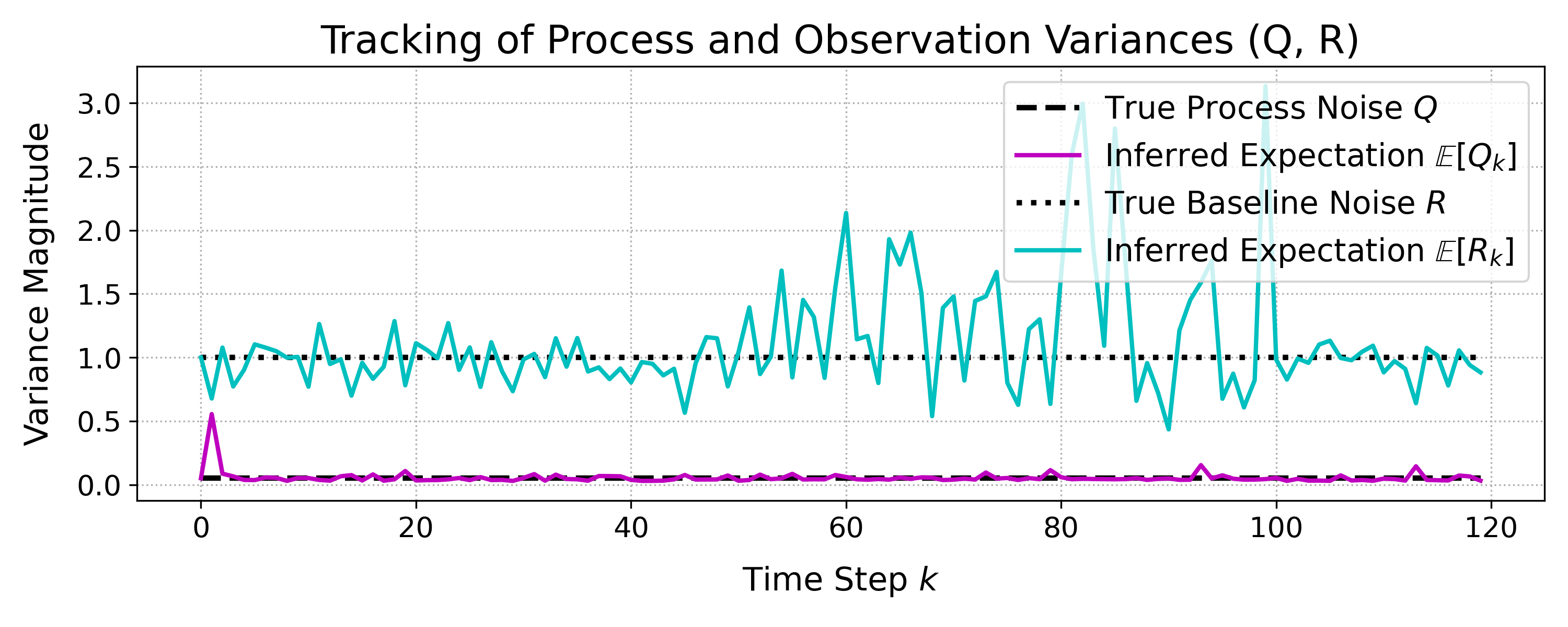}\\
    \vspace{1.5mm}
   \includegraphics[width=1.0\linewidth]{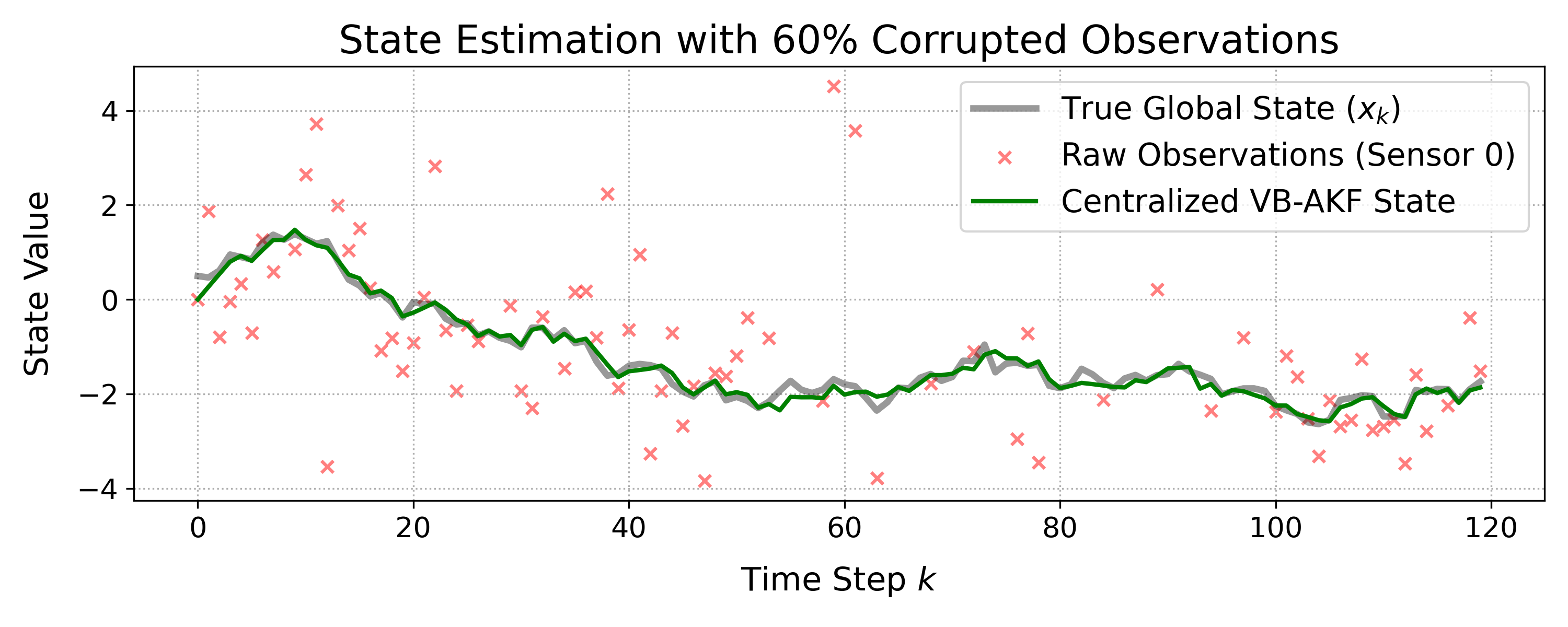} 
   \caption{Performance of the proposed VB-AKF under severe data degradation (60\% packet dropout and 60\% data corruption). Top: Dropout rate inference; Second: Data clean rate identification; Third: Joint inference of process noise $Q_k$ and observation noise $R_k$; Bottom: State estimation via centralized sequential fusion.}
    \label{fig:exp_centralized_fusion}
\end{figure}

As illustrated in Fig.~\ref{fig:exp_centralized_fusion} (Top), the deterministic M-step accurately tracks the global packet dropout rate, which is barely disturbed under severe conditions. Fig.~\ref{fig:exp_centralized_fusion} (Bottom) further validates the robustness of the fusion scheme: despite nearly 84\% of data being invalid, the sequential gated Kalman gain effectively suppresses harmful disturbances. A key observation is the asymmetric robustness among parameters. The inferred process noise $\mathbb{E}[Q_k]$ converges stably to $0.05$ and is nearly immune to corruption and dropouts, whereas the data corruption rate and observation noise $\mathbb{E}[R_k]$ are more sensitive to outliers. This matches the parameter dependence in Fig.~\ref{fig:simple_vi_dep}, as $Q_k$ is less related to $z_{i,k}$ than $R_k$. Even under strong outliers with $E_k=10$, $\mathbb{E}[R_k]$ still fluctuates tightly around $1$.
The soft probabilistic responsibilities $\pi_{i,k}$ prevent outlier residuals from contaminating the inverse-Wishart update, ensuring reliable inference of both $Q_k$ and $R_k$ under extreme data degradation.

\subsection{Sensitivity analysis}\label{sec-4.4}

To clearly analyze the statistical identifiability boundaries of our method, we perform three ablation studies based on the centralized sensor array. By separately adjusting the baseline observation noise \(R_k\), anomaly intensity \(E_k\), and process noise \(Q_k\), we evaluate the RMSE of the inferred corruption rate \(1-\beta_k\).

\begin{figure}[htbp]

    \centering
    \subfigure[Impact of baseline observation noise ($R_k$)]{
        \includegraphics[width=0.95\linewidth]{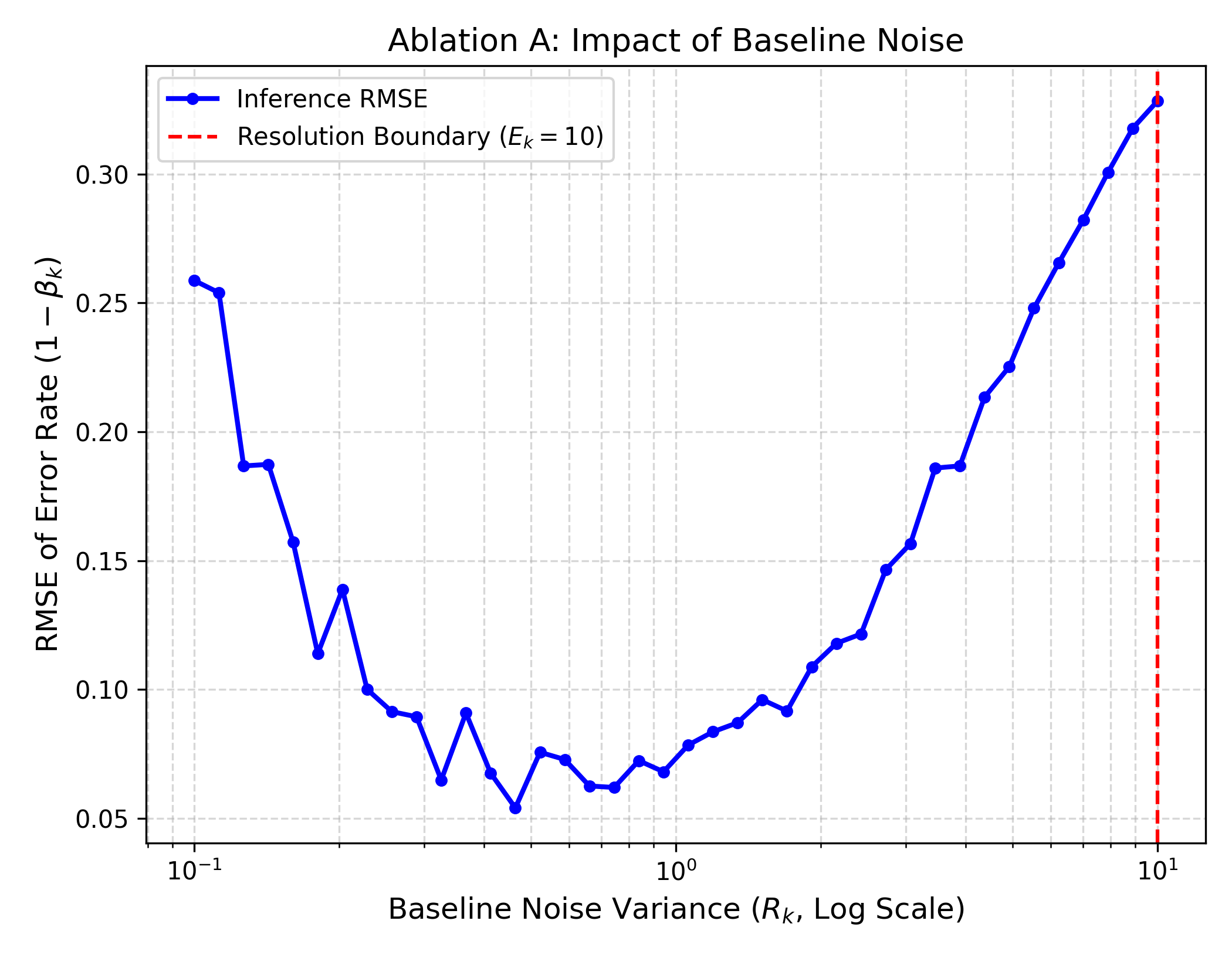}
        \label{fig:ablation_a}
    }
    \vspace{-2mm}
    \subfigure[Impact of anomaly intensity ($E_k$)]{
        \includegraphics[width=0.95\linewidth]{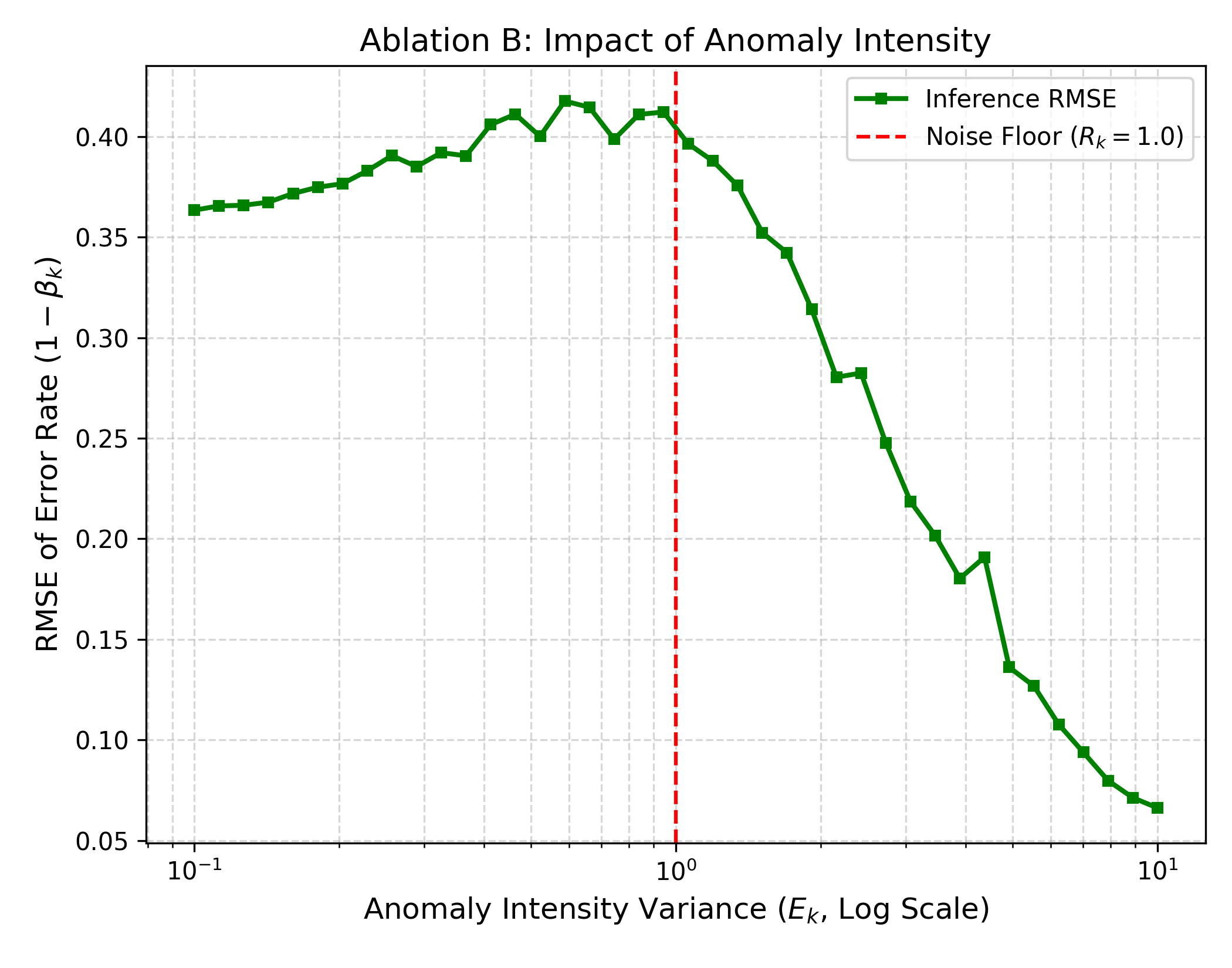}
        \label{fig:ablation_b}
    }
    \vspace{-2mm}
    \subfigure[Impact of process noise ($Q_k$)]{
        \includegraphics[width=0.95\linewidth]{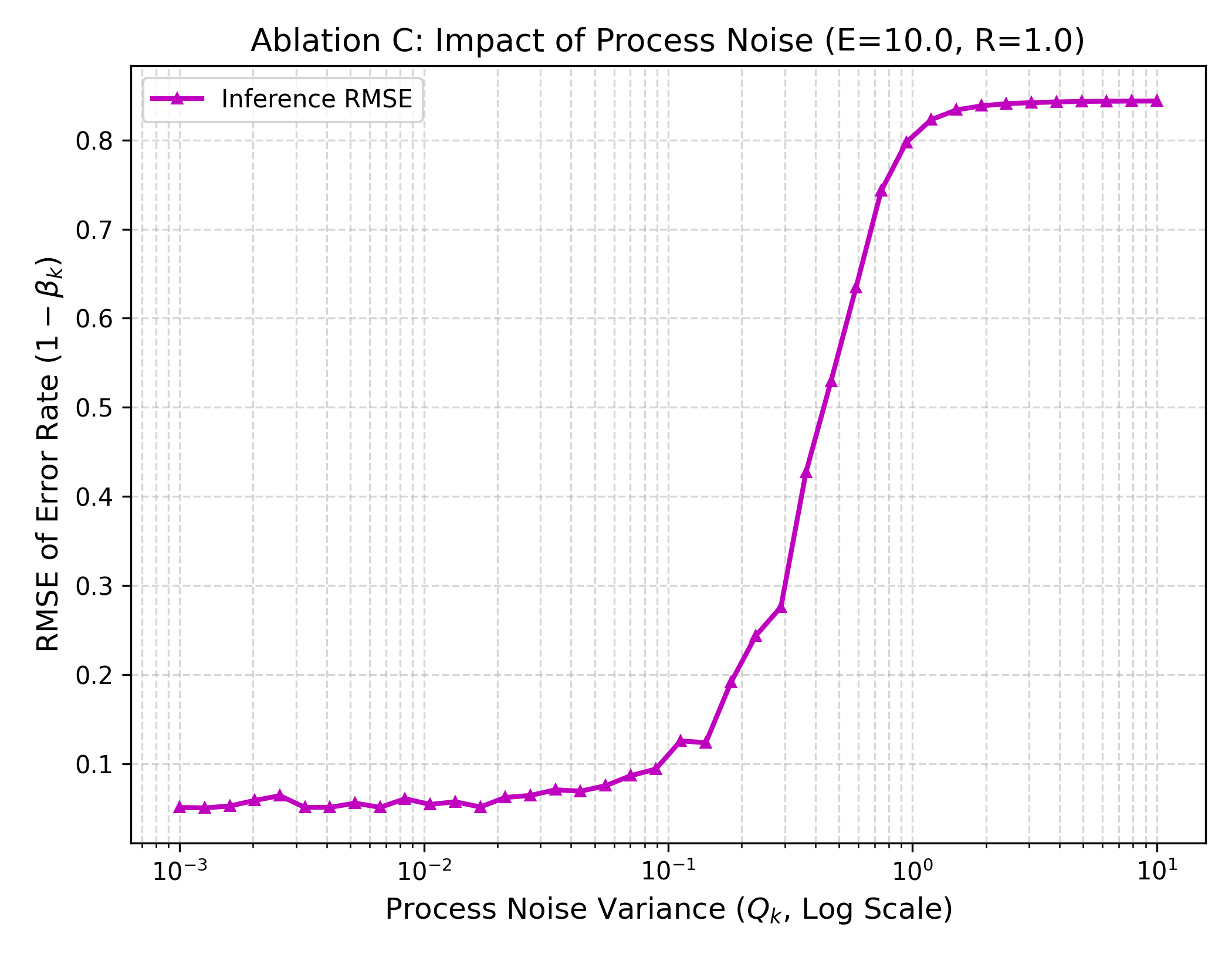}
        \label{fig:ablation_c}
    }
    \caption{Sensitivity analysis and statistical identifiability of corruption rate inference. (a) Identifiability degradation under strong baseline observation noise. (b) Exponential error convergence beyond the anomaly intensity threshold. (c) Identifiability degradation under intense process noise variations.}
    \label{fig:ablation_studies}
\end{figure}

Ablation A: Impact of baseline noise ($R_k$). We vary $R_k$ while fixing the anomaly intensity $E_k=10$ and process noise $Q_k=0.05$. As shown in Fig.~\ref{fig:ablation_a}, the inference RMSE exhibits a clear V-shaped characteristic. When $R_k$ approaches the anomaly intensity $E_k$, the anomalies are severely masked by the background noise, resulting in a substantial loss of statistical identifiability. Conversely, an overly small $R_k$ amplifies the sensitivity to small prediction errors, which easily trigger false-positive classifications due to the narrow confidence interval.The best inference performance is achieved in the intermediate region where the signal-to-anomaly contrast is maximized.

Ablation B: Impact of anomaly intensity ($E_k$). We investigate the inference accuracy under varying anomaly intensities with a fixed baseline noise $R_k=1$. As shown in Fig.~\ref{fig:ablation_b}, when $E_k \leq R_k$, the clean and corrupted Gaussian components overlap significantly, making them statistically indistinguishable. However, once $E_k$ exceeds the baseline noise level, the statistical separation between components increases sharply, and the RMSE decays exponentially toward its theoretical lower bound.

Ablation C: Impact of process noise ($Q_k$). Finally, we examine our proposed algorithm's sensitivity to underlying process variations $Q_k$ with fixed $E_k=10$ and $R_k=1$. The proposed framework maintains robust inference performance for stable targets ($Q_k < 10^{-1}$). However, intense process noise introduces substantial predictive uncertainty $P_{k|k-1}$ into state transitions. Since this uncertainty is incorporated into $\pi_{i,k}$ \eqref{eq:pi_softmax} in through trace expansion \eqref{eq:Delta1}, violent state maneuvers become statistically indistinguishable from sensor corruption. This predictive uncertainty confounding leads to severe overestimation of the corruption rate, which defines the theoretical upper bound of target agility that the dual-mask inference architecture can reliably handle.

    \section{Conclusion}\label{sec-5}

In this paper, a novel variational Bayesian adaptive Kalman filter (VB-AKF) is proposed for state estimation in the presence of simultaneous intermittent packet dropouts and corrupted observations. Different from existing robust adaptive Kalman filtering methods that only address incomplete data or outliers separately, the proposed approach introduces a dual-mask generative model based on two independent Bernoulli random variables, which explicitly characterizes both data loss and measurement corruption. Meanwhile, the VB-AKF framework integrates multiple concurrent observations into the adaptive filtering structure, which significantly improves statistical identifiability and enables both state estimation and parameter identification to asymptotically approach the theoretical optimal lower bound as the number of sensors increases. Within the variational mean-field inference, an inference isolation mechanism and a sequential gated fusion scheme are developed to suppress outlier-induced variance inflation and guarantee strong robustness against severe data anomalies. The effectiveness and superiority of the proposed method are verified through extensive numerical experiments under extreme dual-failure scenarios and rigorous ablation studies.

Future work will focus on two aspects: 1) Modeling: refining the missing measurement model to better accommodate complex real-world engineering scenarios; 2) Theoretical analysis: investigating the coupling relationships among the process noise covariance \(Q_k\), baseline observation noise covariance \(R_k\), and anomaly intensity \(E_k\), as well as their joint impacts on corruption rate inference accuracy. As observed in Section~\ref{sec-4.4}, these three covariance terms interact non-trivially in determining inference performance, which appears to be closely related to controllability and observability arguments in classical linear filtering theory.

    \bibliographystyle{ieeetr}
    \bibliography{references}
        
\end{document}